\pdfoutput=1
\documentclass[sigconf,screen]{acmart}
\renewcommand\footnotetextcopyrightpermission[1]{} 
\AtBeginDocument{%
  \providecommand\BibTeX{{%
    \normalfont B\kern-0.5em{\scshape i\kern-0.25em b}\kern-0.8em\TeX}}}

\usepackage{CJKutf8}
\usepackage{makecell}
\usepackage{arydshln}
\usepackage{multirow}

\begin{document}

\begin{CJK}{UTF8}{gbsn}

\title{Retrieve-Rewrite-Answer: A KG-to-Text Enhanced LLMs Framework for Knowledge Graph Question Answering}

\author{Yike Wu}
\authornote{Both authors contributed equally to this research.}
\email{yike.wu@seu.edu.cn}
\affiliation{%
    \institution{Southeast University}
    \city{Nanjing}
    \state{Jiangsu}
    \country{China}
    }

\author{Nan Hu}
\authornotemark[1]
\email{nanhu@seu.edu.cn}
\affiliation{%
    \institution{Southeast University}
    \city{Nanjing}
    \state{Jiangsu}
    \country{China}
    }

\author{Sheng Bi}
\email{bisheng@seu.edu.cn}
\affiliation{%
    \institution{Southeast University}
    \city{Nanjing}
    \state{Jiangsu}
    \country{China}
    }

\author{Guilin Qi}
\email{gqi@seu.edu.cn}
\affiliation{%
    \institution{Southeast University}
    \city{Nanjing}
    \state{Jiangsu}
    \country{China}
    }

\author{Jie Ren}
\email{renjie@zhejianglab.com}
\affiliation{%
    \institution{Zhejiang Lab}
    \city{Hangzhou}
    \state{Zhejiang}
    \country{China}
    }

\author{Anhuan Xie}
\email{xieanhuan@zhejianglab.com}
\affiliation{%
    \institution{Zhejiang Lab}
    \city{Hangzhou}
    \state{Zhejiang}
    \country{China}
    }

\author{Wei Song}
\authornote{Corresponding author}
\email{weisong@zhejianglab.com}
\affiliation{%
    \institution{Zhejiang Lab}
    \city{Hangzhou}
    \state{Zhejiang}
    \country{China}
    }

\renewcommand{\shortauthors}{Wu and Hu, et al.}
\begin{abstract}

Despite their competitive performance on knowledge-intensive tasks, large language models (LLMs) still have limitations in memorizing all world knowledge especially long tail knowledge. In this paper, we study the KG-augmented language model approach for solving the knowledge graph question answering (KGQA) task that requires rich world knowledge. Existing work has shown that retrieving KG knowledge to enhance LLMs prompting can significantly improve LLMs performance in KGQA. However, their approaches lack a well-formed verbalization of KG knowledge, i.e., they ignore the gap between KG representations and textual representations. To this end, we propose an answer-sensitive KG-to-Text approach that can transform KG knowledge into well-textualized statements most informative for KGQA. Based on this approach, we propose a KG-to-Text enhanced LLMs framework for solving the KGQA task. Experiments on several KGQA benchmarks show that the proposed KG-to-Text augmented LLMs approach outperforms previous KG-augmented LLMs approaches regarding answer accuracy and usefulness of knowledge statements.\footnote{Our code is available at https://github.com/wuyike2000/Retrieve-Rewrite-Answer} 

\end{abstract}



\maketitle

\section{Introduction}
Large language models (LLMs) are becoming increasingly popular in natural language processing for their superior competence in various applications. Although LLMs demonstrate remarkable capabilities in zero-shot scenarios, their performances on several knowledge-intensive tasks are insufficiently satisfactory \cite{DBLP:conf/acl/MallenAZDKH23}. This reveals the enormous parameters in LLMs cannot store all the world's knowledge. Several researches indicate that LLMs still suffer from issues like hallucinations and factual inaccuracy when answering questions \cite{DBLP:conf/emnlp/RohrbachHBDS18, DBLP:journals/csur/JiLFYSXIBMF23}. Specifically, LLMs perform inadequately in the knowledge-intensive task KGQA \cite{hu2023empirical,DBLP:journals/corr/abs-2303-07992}.

To solve this problem, recent work attempts to enhance LLMs with external knowledge. A line of work \cite{DBLP:journals/corr/abs-2304-06975,DBLP:journals/corr/abs-2305-15062, DBLP:journals/corr/abs-2307-14334,DBLP:journals/corr/abs-2302-09432} involves continual pre-training LLMs on extensive corpora. Nevertheless, this method requires a significant amount of textual data, computational resources, and time investment. Some previous work has attempted to explicitly enhance the LLMs' performance on downstream tasks by leveraging external knowledge, such as knowledge graphs and web contents \cite{DBLP:conf/acl/LoganLPGS19, DBLP:conf/emnlp/0010HLHWHC22, komeili-etal-2022-internet}. This approach is employed to address the model's deficiency in factual knowledge. Inspired by this, other work \cite{sen-etal-2023-knowledge,DBLP:journals/corr/abs-2306-04136} constructs knowledge-augmented prompt by prepending question-related factual information to the question to enrich the knowledge of LLMs in a more direct fashion. Although this approach proves to be successful and cost-effective, it ignores the importance of knowledge representation. 

\begin{figure}[h]
    \centering
    \includegraphics[width=\linewidth]{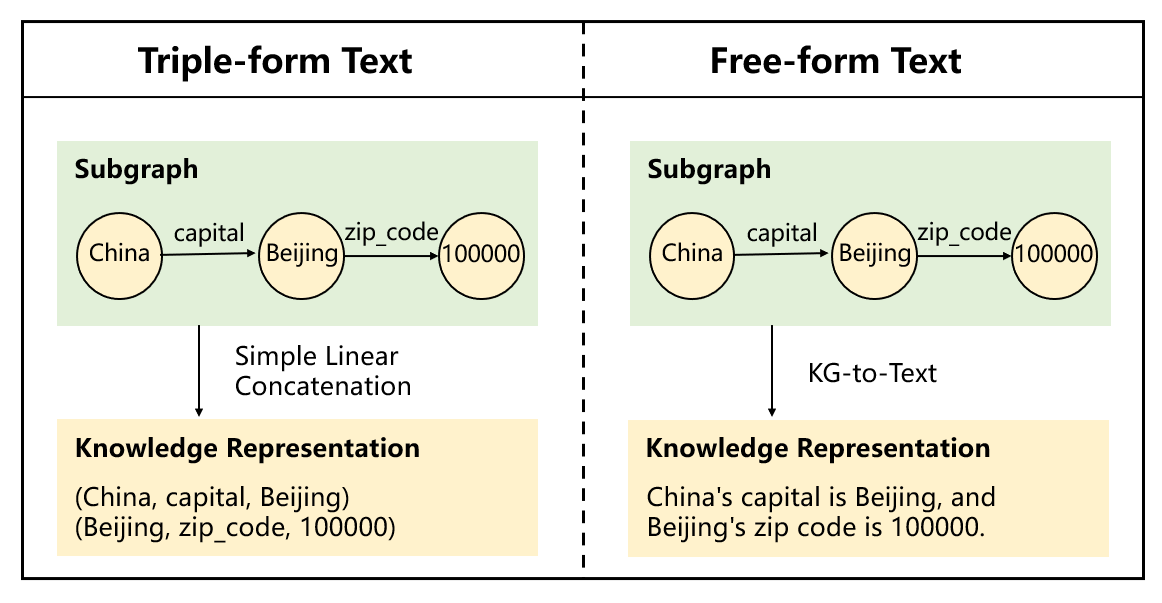}
    \caption{Two forms of knowledge representation: triple-form text and free-form text. Triple-form text transforms subgraph to text via simple linear concatenation while free-form text adopts the KG-to-Text method to generate a semantically coherent textual description.}
    \label{Figure 1}
\end{figure}

\begin{figure}[h]
    \centering
    \includegraphics[width=\linewidth]{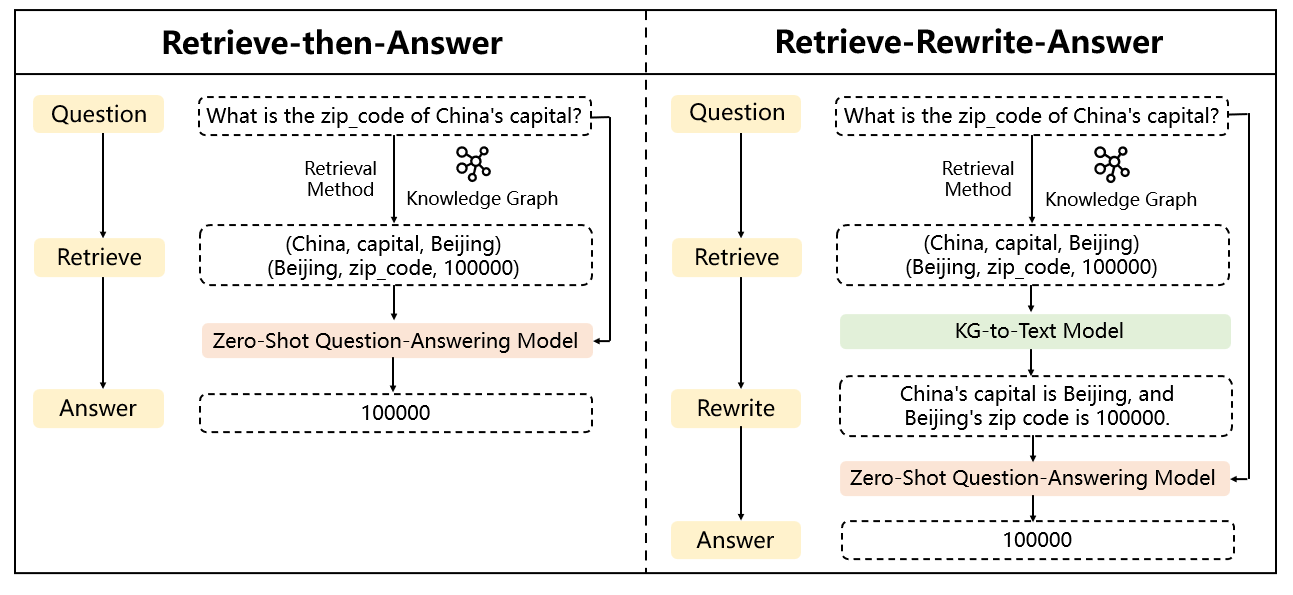}
    \caption{Comparison between Retrieve-then-Answer framework and our proposed Retrieve-Rewrite-Answer framework. We use a rewriter module to transform the retrieved subgraphs into textual descriptions.}
    \label{Figure 2}
\end{figure}

In this paper, we summarize two knowledge representation formats employed in previous work: triple-form text and free-form text. As shown in Figure \ref{Figure 1}, triple-form text involves a simple linear concatenation of triples, transforming them into structured text. Free-form text converts triples into semantically coherent textual descriptions based on the KG-to-Text method. Further, we propose a KG-to-Text enhanced framework, Retrieve-Rewrite-Answer to strengthen the performance of LLMs on KGQA. As shown in Figure \ref{Figure 2}, compared with previous work that answers questions in a Retrieve-then-Answer fashion, our framework adopts a Rewriter module to transform the retrieved triples into textual descriptions. The core of this framework lies in the task-driven KG-to-Text method. Our designed method is answer-sensitive and can transform question-related triples into textual knowledge which is most informative to KGQA. Compared with previous work that simply adopts an off-the-shelf KG-to-Text model, we fine-tune open-source LLMs on the KG-to-Text corpus to generate knowledge descriptions beneficial to KGQA. However, the major challenge is the lack of KG-to-Text annotated data in existing KGQA benchmarks. Therefore, we propose an automatic corpus generation method for generating high-quality graph-text pairs based on the feedback of LLMs.

The contributions of this paper are summarized as follows:
\begin{itemize}

\item We propose Retrieve-Rewrite-Answer, a KG-to-Text enhanced LLMs framework for KGQA. In comparison to previous KG-augmented LLMs frameworks, the most significant innovation in our framework lies in the rewrite module. This module uses fine-tuned LLMs as KG-to-Text models to convert retrieved subgraphs into well-textualized statements most informative for KGQA.

\item To address the issue of scarcity of KG-to-Text corpus annotations, we devise an automatic KG-to-Text corpus generation method. We extract question-related subgraphs and utilize ChatGPT as a tool for corpus generation. Based on the feedback from question-answering LLMs, we generate answer-sensitive knowledge descriptions for the construction of KG-to-Text labeled data. We fine-tune several open-source LLMs on the generated corpus and investigate the impact of textual knowledge generated by different LLMs on KGQA.

\item We evaluate our framework on four KGQA benchmarks. Experimental results show that our framework outperforms previous KG-augmented methods across several LLMs, which demonstrates its effectiveness. Besides, we investigate the influence of different knowledge representation formats on KGQA and prove that the knowledge produced by our framework is most beneficial. 

\end{itemize}

\section{PRELIMINARIES}
\textbf{Knowledge Graph} (KG) is a collection of triples $(s,r,o)$ composed of subject $s$, relation $r$ and object $o$, denoted by $G = \{(s, r, o)|s, o \in E, r \in R\}$, where $E$ and $R$ denote the entity set and relation set.

\textbf{KG-to-Text} is a natural language generation technique based on KG. Given a subgraph $G' = \{(s, r, o)|s, o \in E, r \in R\}$ from KG $G$, the objective of KG-to-Text is to generate a text sequence $X=(x_1, x_2, ..., x_n)$ that is semantically coherent with subgraph $G'$.

\textbf{Knowledge Graph Question Answering} (KGQA) is the task of answering natural language questions based on a set of facts over KGs. Given a question $q$ and a topic entity $e_h$, the task is to generate an answer $a$ that can correctly respond to the question.

\section{RELATED WORK}

\subsection{KG-Augmented LLM for KGQA}
Despite LLMs' pre-training on massive corpora, they still suffer from hallucinations, factual inaccuracy, and outdated knowledge when it comes to KGQA. Recent efforts have aimed to harness KGs to enhance the capabilities of LLMs on KGQA \cite{sen-etal-2023-knowledge,DBLP:journals/corr/abs-2306-04136}. In these studies, question-related triples are extracted from KG and transformed into textual format using techniques such as linear verbalization or off-the-shelf KG-to-Text models. The textual representation of triples and the question are transformed into the knowledge-augmented prompt via predefined templates. The prompt is then fed into the question-answering LLM to produce more reliable answers.

While these studies demonstrate the effectiveness of this strategy, they ignore the impact of the knowledge representation format on the performance of LLMs on KGQA. In this study, we devise a KG-to-Text corpus generation method to produce high-quality annotations. Then we utilize LLMs fine-tuned on the corpus to transform question-related subgraphs into knowledge text, which can further elevate the performance of LLMs on KGQA.

\subsection{KG-to-Text}
Recent studies in KG-to-Text can be classified into two main approaches: 1) based on graph neural networks (GNNs) \cite{DBLP:conf/iclr/VelickovicCCRLB18}: In order to preserve the structural information of subgraphs during encoding, researchers focus on designing more complex encoders to obtain enhanced subgraph representation. Some work has developed graph-structured encoders based on GNNs \cite{DBLP:conf/inlg/MarcheggianiP18, DBLP:journals/tacl/GuoZTL19} due to the robust capabilities of GNNs in encoding graph-structured data. However, GNNs are restricted by local processing nature, as they iteratively compute node representations based on neighboring nodes' features. To overcome this limitation, alternative efforts have employed the Transformer-based architecture \cite{DBLP:conf/nips/VaswaniSPUJGKP17} in designing encoders \cite{DBLP:conf/naacl/Koncel-Kedziorski19,DBLP:conf/emnlp/ZhuLZQZZ19,DBLP:conf/aaai/CaiL20}. This approach allows for the establishment of connections between any two nodes within the graph, thereby addressing the limitations of GNNs. 2) based on pre-trained language models (PLMs): With the development of large-scale generative PLMs such as BART \cite{DBLP:conf/acl/LewisLGGMLSZ20}, T5 \cite{DBLP:journals/jmlr/RaffelSRLNMZLL20} and GPT \cite{radfordlanguage}, recent work applies these models to KG-to-Text, modeling it as an end-to-end generation task \cite{DBLP:conf/acl/KeJRCWSZH21,DBLP:conf/emnlp/HanS22,ribeiro-etal-2021-investigating}. These studies involve modifications to the model architecture and the introduction of pre-training tasks to enhance the ability to extract structural information. In this work, we follow the second approach and fine-tune open-source LLMs on the KG-to-Text corpus.

\section{Methods}
\subsection{Retrieve-Rewrite-Answer}
\begin{figure}
    \centering
    \includegraphics[width=\linewidth]{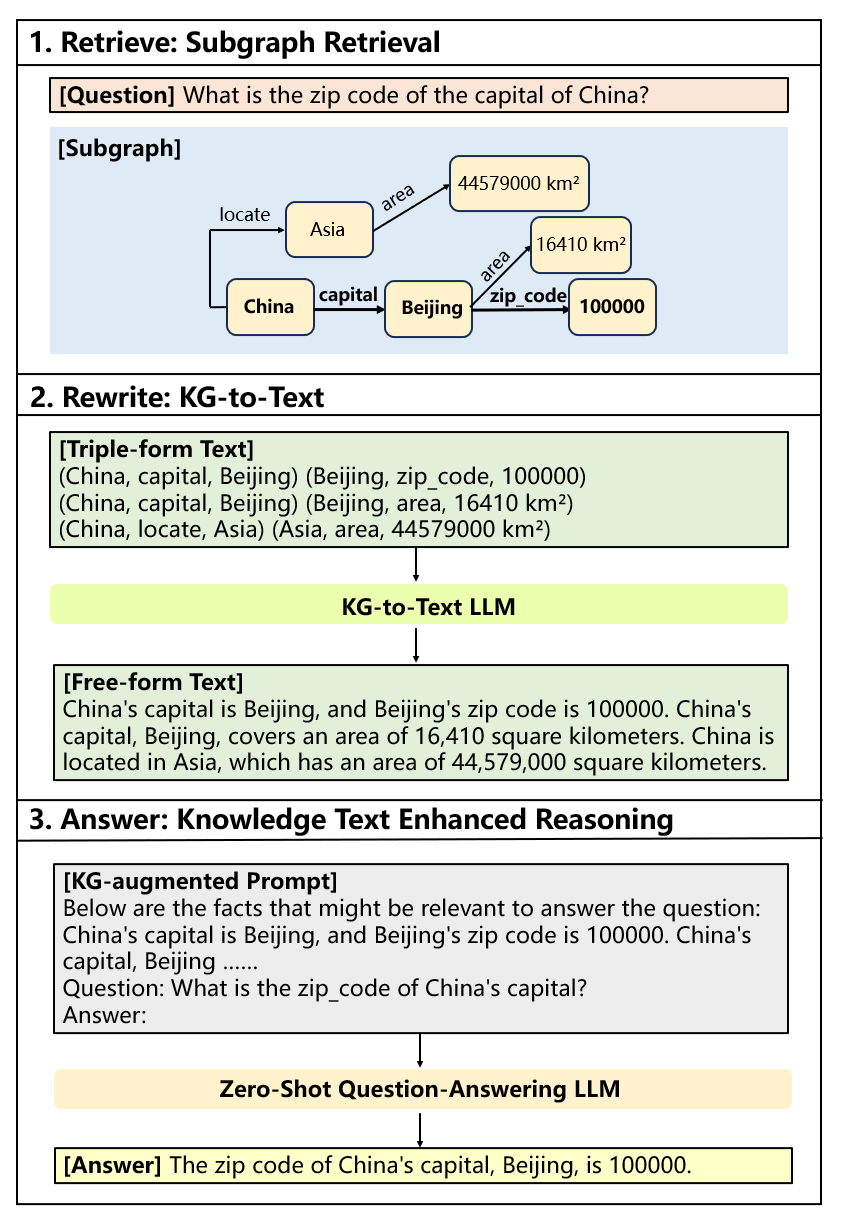}
    \caption{Our proposed framework, Retrieve-Rewrite-Answer has three steps: subgraph retrieval, KG-to-Text, and knowledge text enhanced reasoning.}
    \label{Figure 3}
\end{figure}

As shown in Figure \ref{Figure 3}, our proposed Retrieve-Rewrite-Answer framework contains three steps: subgraph retrieval, KG-to-Text, and knowledge text enhanced reasoning.

\subsubsection{Retrieve: Subgraph Retrieval}
\begin{figure}
    \centering
    \includegraphics[width=\linewidth]{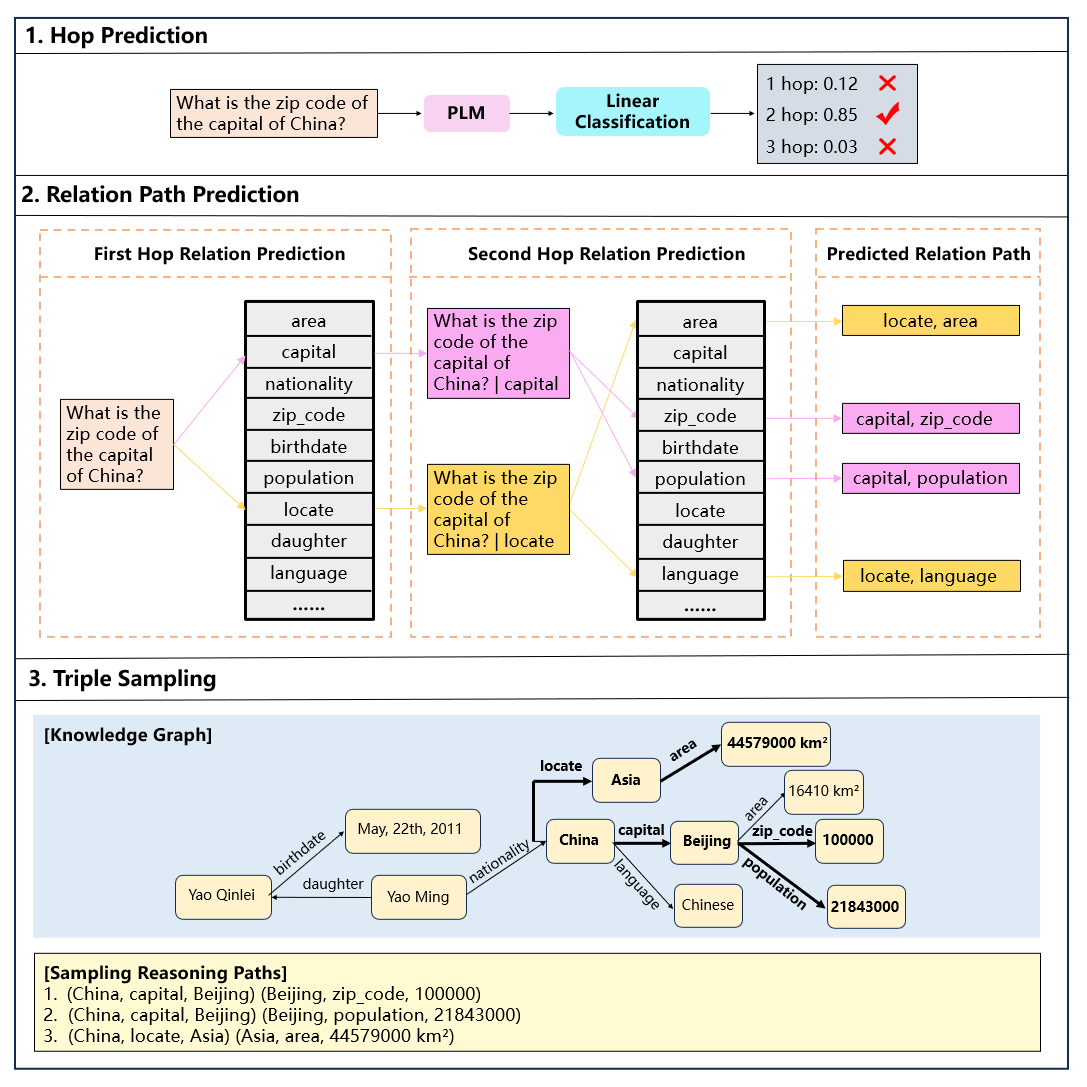}
    \caption{Our retrieve module consists of three steps: hop prediction, relation path prediction, and triple sampling.}
    \label{Figure 4}
\end{figure}

Our retrieve module consists of three steps: hop prediction, relation path prediction, and triple sampling, as illustrated in Figure \ref{Figure 4}.

\textbf{Hop Prediction}. The aim of this step is to predict the hop number of the question, which is used to predict the relation path in the next step. We model the hop prediction as a classification task based on PLMs. Given the question $q$, we employ PLMs to encode the question $q$ and obtain the vector representation $q_v$:
\begin{equation}
q_v=PLM(q)
\end{equation}
The representation $q_v$ is then fed into the linear classification layer to predict the probability distribution $D'_h$ of potential hop numbers $h_1$, $h_2$, ..., $h_H$:
\begin{equation}
D'_h=[d'_{h_{1}}, d'_{h_2}, ..., d'_{h_H}]=Linear(q_v)
\end{equation}
where $d'_{h_{c}}$ is the probability of the hop number $h_c$ given the question representation $q_v$:
\begin{equation}
d'_{h_{c}}=P(h_c|q_v),\quad c=1,2, ..., H
\end{equation}
The hop number $h$ with the highest probability is selected as the predicted result.
\begin{equation}
h=\arg\max_{h_c}d'_{h_c},\quad c=1,2, ..., H
\end{equation}
During training, ground truth distribution $D_h$ is represented as a one-hot vector with a probability of 1 for the true hop number $h_{gold}$ and probabilities of 0 for other hop numbers:
\begin{equation}
D_h=[d_{h_{1}}, d_{h_2}, ..., d_{h_H}]
\end{equation}
\begin{equation}
d_{h_c}=
\begin{cases}
1, & h_c=h_{gold}\\
0, & h_c \neq h_{gold}
\end{cases}
\quad c=1,2, ..., H
\end{equation}
The predicted distribution $D'_h$ is penalized for being different from the ground truth distribution $D_h$ using cross entropy loss $L_{CE}$:
\begin{equation}
L_{CE}=-D_h\log{D'_h}=-\sum_{c=1}^{H}d_{h_c}\log{d'_{h_c}}
\end{equation}
$L_{CE}$ is utilized to update the model's parameters.

\textbf{Relation Path Prediction}.
Given the question $q$ and the predicted hop number $h$, we perform $h$-step prediction, with each step corresponding to one hop relation. At step $t$, we predict the $t$-th hop relation based on the predicted $(t-1)$ hop relation paths and the question $q$ via PLMs as a classification task. Specifically, for each predicted relation path, we sample $K$ candidate relations for the next step. In step 1, we encode the question $q$ into a vector representation $q_v$. This vector is then passed through a linear classification layer to compute the distribution $D'_{r,1}$ of $R$ relations in KG:
\begin{equation}
D'_{r,1}=[d'_{r_1}, d'_{r_2}, ..., d'_{r_R}]=Linear(q_v)
\end{equation}
where $d'_{r_c}$ is the probability of relation $r_c$ given the question representation $q_v$:
\begin{equation}
d'_{r_c}=P(r_c|q_v),\quad c=1,2, ..., R
\end{equation}
We select top $K$ relations with the highest probabilities as one hop relation paths $p_1$.

In the following step $t$ ($t>1$), the relation path $p_{t-1,i}$ in the $(t-1)$ hop relation paths $p_{t-1}$ can be represented as:
\begin{equation}
p_{t-1,i}=r_{i,1} | r_{i,2} | ... | r_{i,t-1}, \quad i=1,2, ..., K^{t-1}
\end{equation}
We concatenate the question $q$ and the relation path $p_{t-1,i}$ with ``|'' as the input sequence $Q_{t}$:
\begin{equation}
Q_{t}=q | r_{i,1} | r_{i,2} | ... | r_{i,t-1}
\end{equation}
$Q_{t}$ is encoded into vector representation $Q_{t,v}$ via PLMs and passes through a linear classification layer to calculate the relation distribution $D'_{r,t}$ across $R$ relations in KG:
\begin{equation}
Q_{t,v}=PLM(Q_t)
\end{equation}
\begin{equation}
D'_{r,t}=[d'_{r_1}, d'_{r_2}, ..., d'_{r_R}]=Linear(Q_{t,v})
\end{equation}
where $d'_{r_c}$ is the probability of relation $r_c$ given the input sequence representation $Q_{t,v}$:
\begin{equation}
d'_{r_c}=P(r_c|Q_{t,v}),\quad c=1,2, ..., R
\end{equation}
We retain top $K$ relations with the highest probabilities as the $t$-th hop relations for the $(t-1)$ hop relation path $p_{t-1,i}$. 

After $h$-step prediction, we can obtain $K^h$ relation paths. The score of the relation path $p_{t,i}$ is the product of the probabilities of all relations in the path:
\begin{equation}
Score(p_{t,i})=Score(r_{i,1} | r_{i,2} | ... | r_{i,t})=\prod_{l=1}^{t} d'_{r_{i,l}},\; i=1,2, ..., K^h
\end{equation}

We employ a training method similar to hop prediction. For the question $q$ and the ground truth relation path, we concatenate question $q$ and $(t-1)$ hop relation path as the inputs $Q_{t-1}$. We encode $Q_{t-1}$ into vector representation and feed it into the linear classification layer to obtain the $t$-th hop relation distribution $D'_{r,t}$. The ground truth distribution $D_{r,t}$ is represented as a one-hot vector with a probability of 1 for the true $t$-th hop relation and a probability of 0 for other relations. We use the cross-entropy loss for parameter optimization.

\textbf{Triple Sampling}.
We arrange the predicted relation paths in descending order of scores and sequentially sample reasoning paths composed of triples from KG until the number of reasoning paths reaches $M$. These reasoning paths are utilized as the relevant knowledge to augment LLMs on KGQA.

\subsubsection{Rewrite: KG-to-Text} The core of our rewrite module is to transform structured triples into free-form text based on a KG-to-Text model. We start by training an open-source LLM based on question-related graph-text pairs. Given the graph $G$ and corresponding free-form text $y$. We verbalize the triples in graph $G$ into triple-form text $x$ by concatenating the subject, relation, and object. Then, we transform triple-form text $x$ to graph-to-text transformation prompt $p1$ via a template $T1$:``\textit{Your task is to transform a knowledge graph to a sentence or multiple sentences. The knowledge graph is:} \{triple-form text $x$\}. \textit{The sentence is:}''. We take prompt $p1$ and free-form text $y$ as the input and output respectively and adopt the teacher forcing strategy in training. Formally, given the prompt $p1$, ground truth output sequence $y=[y_1, y_2, ..., y_T]$ and model vocabulary $[v_1, v_2, ..., v_V]$, the model predicts the probability distribution $D'_{v,t}$ of tokens at step $t$ based on prompt $p1$ and previous $(t-1)$ steps correct tokens $y_1, y_2, ..., y_{t-1}$:
\begin{equation}
D'_{v,t}=[d'_{v_1}, d'_{v_2}, ..., d'_{v_V}]
\end{equation}
where $d'_{v_c}$ is the probability of $v_c$ given $p1$ and $y_1, y_2, ..., y_{t-1}$: 
\begin{equation}
d'_{v_c}=P(v_c|p1, y_1, y_2, ..., y_{t-1}), \quad c=1,2, ..., V
\end{equation}
Ground truth distribution $D_{v,t}$ is a one-hot vector with a probability of 1 for the true next token $y_t$ and probabilities of 0 for other tokens:
\begin{equation}
D_{v,t}=[d_{v_1}, d_{v_2}, ..., d_{v_V}]
\end{equation}
\begin{equation}
d_{v_c}=
\begin{cases}
1, & v_c=y_t\\
0, & v_c \neq y_t
\end{cases}
\quad c=1,2, ..., V
\end{equation}

The cross entropy loss $L_{CE}$ is used to update the parameters:
\begin{equation}
J_t=-D_{v,t}\log{D'_{v,t}}=-\sum_{c=1}^{V}d_{v,c}\log{d'_{v,c}}
\end{equation}
\begin{equation}
L_{CE}=\frac{1}{T}\sum_{t=1}^{T}J_t
\end{equation}

While answering questions, each reasoning path is first linearized into triple-form text and then transformed into a prompt via template $T1$. The prompt is fed into the fine-tuned LLMs to obtain the corresponding textual description. These descriptions are consolidated into a paragraph as the question-relevant knowledge to enhance the performance of the LLMs.

\subsubsection{Answer: Knowledge Text Enhanced Reasoning} To integrate the generated knowledge $y$ with the question $q$, we devise a template $T2$:``\textit{Below are the facts that might be relevant to answer the question:} \{free-form text $y$\} Question:  \{question $q$\} Answer:''. We map the free-form text $y$ and the question $q$ to KG-augmented prompt $p2$ using template $T2$. Then we input prompt $p2$ into the question-answering model and collect output as the predicted answer $a$.

\subsection{Corpus Generation}

\begin{figure}
    \centering
    \includegraphics[width=\linewidth]{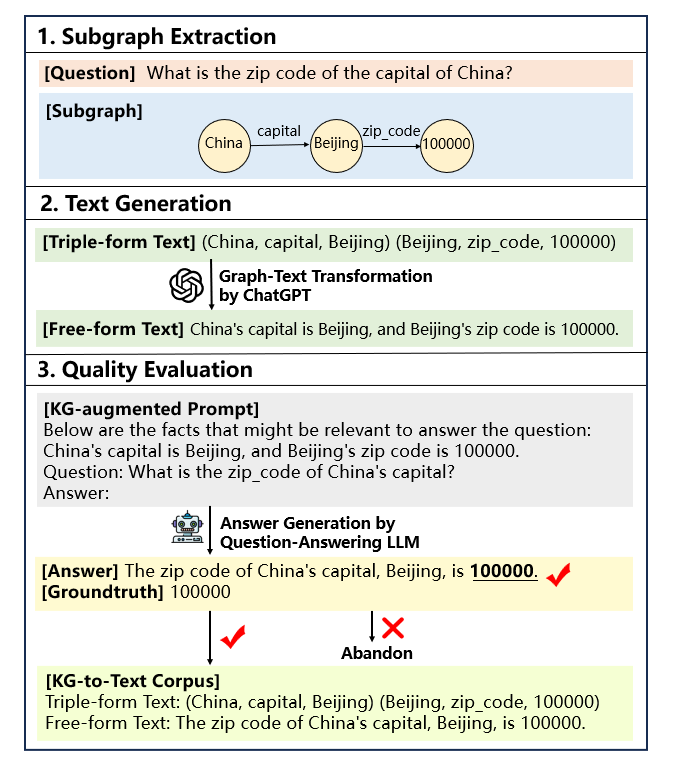}
    \caption{Our designed KG-to-Text corpus generation method has three steps: subgraph extraction, text generation, and quality evaluation.}
    \label{Figure 5}
\end{figure}

Existing KGQA benchmarks do not provide graph-text pairs for the question-answering task. Therefore, we design a KGQA task-driven corpus generation method. Considering ChatGPT's powerful natural language understanding and generation capabilities, inspired by \cite{DBLP:journals/corr/abs-2305-06147,dai2023auggpt, ostyakova2023linguistic}, we adopt ChatGPT as the corpus generator. As shown in Figure \ref{Figure 5}, this process contains three steps: subgraph extraction, text generation, and quality evaluation.

\subsubsection{Subgraph Extraction} 
For KGQA benchmarks that provide relation paths or reasoning triples (e.g. MetaQA \cite{DBLP:conf/aaai/ZhangDKSS18}, PathQuestion \cite{DBLP:conf/coling/ZhouHZ18}), we obtain subgraphs by querying KG based on annotations. For benchmarks annotated with SPARQL (e.g. WebQuestionSP \cite{DBLP:conf/acl/YihRMCS16}, ComplexWebQuestion \cite{DBLP:conf/naacl/TalmorB18}), we modify SPARQL query to retrieve intermediate entities and construct question-related subgraphs.

\subsubsection{Text Generation} 
Given the question $q$, we first verbalize the related subgraph $G$ into triple-form text $x$ by concatenating the subject, relation, and object. Then, template $T1$ is adopted to transform triple-form text $x$ to graph-to-text transformation prompt $p1$. Finally, we input prompt $p1$ into ChatGPT to obtain the corresponding free-form text $y$.

\subsubsection{Quality Evaluation}
Due to the absence of annotations, common evaluation metrics including BLEU, METEOR, and ROUGE cannot be employed. Considering the purpose of the generated text is to strengthen the performance of LLMs on KGQA, we assess the quality of free-form text $y$ based on the feedback from question-answering models. We map the free-form text $y$ and question $q$ to KG-augmented prompt $p2$ via template $T2$. Then, we feed prompt $p2$ to the question-answering model and get the predicted answer $a$. In light of the fact that LLMs typically provide textual passages as responses instead of single answer entities, we employ hit@1 as a metric to assess the correctness of answer $a$. To put it simply, if answer $a$ contains at least one answer entity, the question is considered answered correctly. In this scenario, we collect the triple-form text $x$ and free-form text $y$ as a graph-text pair.

\section{Experiments}

\begin{table*}[h]
\caption{Experimental results of our proposed framework and baselines on WebQSP/WebQ, where we report hit@1(\%). We implement our KG-to-Text model based on Llama-2-chat (13B). The results of the baselines are copied from the original paper. The best score is emphasized in bold.}
\begin{tabular}{l|ccccc|ccccc}
    \hline
    \multirow{2}{*}{\textbf{Method}}& \multicolumn{5}{c|}{\textbf{WebQSP}} & \multicolumn{5}{c}{\textbf{WebQ}}\\
    \cline{2-11}
     & \textbf{T5$_{(0.8B)}$} & \textbf{T5$_{(3B)}$} &\textbf{T5$_{(11B)}$} &\textbf{T0$_{(3B)}$} & \textbf{T0$_{(11B)}$} &  \textbf{Flan-T5$_{(80M)}$} &\textbf{Flan-T5$_{(3B)}$} &\textbf{Flan-T5$_{(11B)}$} & \textbf{T0$_{(3B)}$} &  \textbf{T0$_{(11B)}$}\\
    \hline
    KAPING & 34.70& 25.41& 24.91& 52.28 & 62.85 & - & - &- & - &-\\
    Sen et al. &- & - &- & - &-&  45.52&55.58& 59.79 & 53.33 & 55.64\\
    \textbf{Ours}& \textbf{69.41} & \textbf{72.11}&\textbf{79.36} &\textbf{74.02} & \textbf{68.55}  &\textbf{59.95} & \textbf{68.92}&\textbf{73.71} &\textbf{74.02} & \textbf{68.55}\\
    \hline
\end{tabular}
\label{Table 1}
\end{table*}

\subsection{Datasets}

\textbf{MetaQA} \cite{DBLP:conf/aaai/ZhangDKSS18} is a large-scale multi-hop KGQA benchmark in the movie domain. It provides a knowledge graph including 135k triples, 43k entities, and 9 relations. It contains more than 400k questions divided into MetaQA 1-hop, MetaQA 2-hop, and MetaQA 3-hop based on the hop number of the question. Each question is annotated with the head entity, answers, and entity categories involved in the reasoning path. In this experiment, we choose MetaQA 2-hop as our benchmark and the``vanilla'' version of the questions, totaling 148,724 questions (118,980 train, 14,872 dev, 14,872 test). We collect the gold relation paths based on the provided entity categories since there is only one type of relation between two categories of entities.

\textbf{WebQuestionsSP (WebQSP)} \cite{DBLP:conf/acl/YihRMCS16} is a smaller scale KGQA benchmark with a larger scale KG. It provides SPARQL queries for WebQuestions \cite{DBLP:conf/emnlp/BerantCFL13} and filters out questions that are not answerable. The remaining 4,737 questions (3,098 train, 1,639 test) are either 1-hop or 2-hop questions with corresponding topic entities, inferential chains, and SPARQL queries. Following \cite{DBLP:conf/acl/SaxenaTT20}, we prune the KG to contain only relations mentioned in the questions and the triples within 2 hops of mentioned entities. The pruned KG includes 1.8 million entities, 627 relations, and 5.7 million triples. 

\textbf{WebQuestions (WebQ)} \cite{DBLP:conf/emnlp/BerantCFL13} collects questions from web pages via Google Suggest API. Following \cite{sen-etal-2023-knowledge}, we use a subset of this benchmark. Our method requires annotations for relation paths or SPARQL queries, which are not provided. Therefore, we choose the questions that are provided with SPARQL queries in WebQSP for training and testing. This results in 4,737 questions (3,098 train, 1,639 test). We use the same KG as WebQSP.

\textbf{ZhejiangQA (ZJQA)} is a Chinese KGQA dataset of 20,491 questions provided by Zhejiang Lab. We divide these questions into trainset, devset, and testset (14,999 train, 2,147 dev, 3,345 test). The questions are either 1-hop or 2-hop questions primarily in the robotic domains. Each question is provided with a head entity, answers, and a gold relation path. It also provides a KG with more than 11k triples, 9k entities, and 39 relations.

\subsection{Large Language Models}
Our proposed framework has two modules based on LLMs: KG-to-Text and question-answering. We use Llama-2 (7B, 13B), Flan-T5 (3B) for KG-to-Text and Llama-2 (7B, 13B), T5 (0.8B, 3B, 11B), Flan-T5 (80M, 3B, 11B), T0 (3B, 11B), ChatGPT for question-answering.

\textbf{Llama-2} \cite{DBLP:journals/corr/abs-2307-09288}
is an updated version of Llama-1 \cite{DBLP:journals/corr/abs-2302-13971}, which is trained on a wide range of public online data sources. Among different variants, we adopt Llama-2-chat (7B, 13B) as our KG-to-Text and question-answering model. For ZJQA, we employ the Chinese version of this model, Chinese-Alpaca-2 (7B, 13B)\footnote{https://github.com/ymcui/Chinese-LLaMA-Alpaca-2}.

\textbf{T5} \cite{DBLP:journals/jmlr/RaffelSRLNMZLL20} is an encoder-decoder model pre-trained on multiple tasks in a text-to-text format. Following \cite{DBLP:journals/corr/abs-2306-04136}, we use the LM-adapted version\footnote{https://github.com/google-research/text-to-text-transfer-transformer/blob/main/released\_checkpoints.md} of T5 as the question-answering model on WebQSP to ensure a fair comparison. 

\textbf{Flan-T5} \cite{DBLP:journals/corr/abs-2210-11416}
is an extension of T5, which is further instruction tuned on a large-scale collection of automatically generated instructions from existing datasets. We use Flan-T5-XL (3B) as KG-to-Text model for MetaQA and Flan-T5-Small (80M), Flan-T5-XL (3B), Flan-T5-XXL (11B) as question-answering models for WebQ. We do not use this model for ZJQA since it does not support Chinese.

\textbf{T0} \cite{DBLP:conf/iclr/SanhWRBSACSRDBX22} is fine-tuned from T5 over a variety of prompts to improve zero-shot generalization performance. We use T0 (3B, 11B) as the question-answering model to compare our proposed framework with previous work on WebQSP and WebQ.  

\textbf{ChatGPT}\footnote{https://openai.com/blog/chatgpt}
is a large language model built on GPT-3.5 developed by OpenAI. It is pre-trained on enormous corpus and human annotations and excels in understanding and generating human-like text. Specifically, we use GPT-3.5 Turbo in this experiment. We cannot fine-tune ChatGPT since it is not yet open source. Therefore, we access it via API\footnote{https://api.openai.com/} and use it as the question-answering model.

\subsection{Baselines}

We compare the proposed framework with previous KG-augmented LLM methods for KGQA on WebQSP and WebQ:

\textbf{KAPING} \cite{DBLP:journals/corr/abs-2306-04136}
verbalizes triples by concatenating subject, relation, and object. During the retrieval process, verbalized triples and questions are embedded into vector space with an off-the-shelf sentence embedding model. The vector similarity between questions and verbalized triples is calculated to retrieve triples as augmented knowledge. The retrieved triples and questions are concatenated in the form of the prompt and fed into the question-answering model to obtain answers. We compare with this baseline on WebQSP.

\textbf{Sen et al.} \cite{sen-etal-2023-knowledge}
improves the retrieval method of KAPING \cite{DBLP:journals/corr/abs-2306-04136}. This work utilizes Rigel \cite{DBLP:conf/emnlp/SaffariOSA21} to predict the distribution over relations for each hop and retrieve triples based on the calculated relation distribution and question entities. The retrieved triples are linear verbalized and concatenated with the question via prompt and input into the question-answering model to generate the answer. We compare with this baseline on WebQ.

\begin{table*}[h]
\caption{Experimental results of different knowledge representation formats on MetaQA and ZJQA, where we report hit@1(\%). The below section of the table is the results of our framework implemented by different KG-to-Text models. The number inside the parentheses behind the LLMs denotes its parameter size. The best score for each question-answering model is emphasized in bold.}
\begin{tabular}{l|ccc|ccc}
    \hline
    \multirow{2}{*}{\textbf{Knowledge Format}}& \multicolumn{3}{c|}{\textbf{MetaQA}} & \multicolumn{3}{c}{\textbf{ZJQA}}\\
    \cline{2-7}
     & \textbf{Llama-2-chat$_{(7B)}$} & \textbf{Llama-2-chat$_{(13B)}$} & \textbf{ChatGPT} &  \textbf{Llama-2-chat$_{(7B)}$} & \textbf{Llama-2-chat$_{(13B)}$} & \textbf{ChatGPT}\\
    \hline
    No Knowledge & 34.31 & 34.08 & 60.58& 14.95 & 15.75& 15.25\\
    Triple Knowledge & 96.98 & 96.78 & 93.19  & 88.67 & 84.48 & 92.02\\
    MTL Knowledge & 86.32 & 87.18 & 80.66 &  - & - & -\\
    MVP Knowledge & 96.13 & 98.04 & 92.91 &  - & - & -\\
    \cline{1-7}
    \textbf{Flan-T5$_{(3B)}$}& 97.60 & 98.78 & \textbf{97.71}  & - & - & -\\
    \textbf{Llama-2-chat$_{(7B)}$}& 97.74 & 98.92 & 97.69  & \textbf{93.09} & \textbf{93.21} & \textbf{92.91}\\
    \textbf{Llama-2-chat$_{(13B)}$}& \textbf{97.77} & \textbf{99.07} & 97.63 & 91.48 & 92.11 & 92.11\\
    \hline
\end{tabular}
\label{Table 2}
\end{table*}

\subsection{Evaluation Metrics}

Following the evaluation settings in previous work on generative KGQA\cite{DBLP:conf/ijcai/YinJLSLL16, DBLP:conf/coling/SenAS22,DBLP:conf/acl/SaxenaKG22}, we use hit@1, which measures whether the generated answer includes at least one answer entity. Both the predicted answers and the answer entities are converted to lowercase to mitigate matching issues arising from letter-case differences.

\subsection{Implementation Details}

\textbf{WebQSP/WebQ}
 We parse the SPARQL query to extract the gold relation path for each question. Bert-base-uncased is used as the classification model for hop prediction and relation path prediction. We modify the SPARQL query and obtain intermediate entities to construct the gold subgraph for each question. All the questions in train split are used for corpus generation. Over 12k graph-text pairs are generated and utilized for supervised fine-tuning. In addition, we exclude 11 test samples without answers.

\textbf{MetaQA} 
We skip the hop prediction step and use the provided ground truth hop number. Bert-base-uncased is used as the classification model for relation path prediction. We randomly sample 17k questions from train split for corpus generation and generate more than 13k graph-text pairs for KG-to-Text fine-tuning.

\textbf{ZJQA}
We use bert-base-chinese for hop prediction and relation path prediction. We randomly sample 14k questions from train split for corpus generation and obtain over 13k KG-to-Text annotations.

For WebQSP and WebQ, we set K=5 (i.e. sample 5 relations as the next possible relations for each predicted relation path) and M=5 (i.e. sample up to 5 reasoning paths for each question) in the subgraph retrieval process. For MetaQA and ZJQA, we set K=3 and M=5. ChatGPT is used as the question-answering model in the corpus generation for all benchmarks. Our framework is implemented using Pytorch\footnote{https://pytorch.org/}, Transformers\footnote{https://huggingface.co/docs/transformers/main/index} and Peft libaries\footnote{https://huggingface.co/docs/peft/index}. For efficient training, all fine-tuning processes adopt LoRA \cite{DBLP:conf/iclr/HuSWALWWC22}. We train KG-to-Text models on 4 NVIDIA Tesla V100 GPUs for 10 epochs with a total batch size of 128 and run inference over KG-to-Text models and question-answering models on 1 NVIDIA Tesla V100 GPUs. We use AdamW optimizer with initialized learning rate 1e-4. The LoRA rank, LoRA alpha, and LoRA dropout are 64, 128, and 0.05.

\subsection{Main Results}

Table \ref{Table 1} shows the overall results of our proposed framework and baselines on WebQSP and WebQ. Our KG-to-Text model is implemented based on Llama-2-chat (13B). For the question-answering model, we choose T5 (0.8B, 3B, 11B), T0 (3B, 11B) and Flan-T5 (80M, 3B, 11B), T0 (3B, 11B) as question-answering models on WebQSP and WebQ respectively. Experimental results demonstrate that our framework outperforms baselines by a large margin across various LLMs. Notably, it exhibits the most significant advantages on T5. This suggests that T5, which is pre-trained solely on textual data, may have limitations in comprehending structured data. This proves that transforming triple-form text into free-form text can enable LLMs to better understand the provided factual knowledge and enhance their capabilities on KGQA.

\subsection{Ablation Study}

We conduct a comparative analysis of different knowledge representation formats and implement our framework utilizing a range of LLMs on MetaQA and ZJQA datasets. This ablation study is performed to investigate the impact of knowledge representation formats and LLMs on KGQA.

Specifically, we use Llama-2-chat (7B, 13B), Flan-T5-XL (3B) as the KG-to-Text model and Llama-2-chat (7B, 13B), ChatGPT as the question-answering model. We compare knowledge generated by our framework with no knowledge, triple knowledge, and knowledge generated by the off-the-shelf KG-to-Text model: 

\textbf{No Knowledge} 
Questions are directly fed into LLMs without additional knowledge. We set this knowledge representation format to explore how much improvement KG-augmented methods can bring to LLMs on KGQA.

\textbf{Triple Knowledge} 
Triple knowledge is the most common strategy used in previous work. We first deduplicate the retrieved triples to mitigate semantic redundancy and then simply verbalize each triple by concatenating the subject, relation, and object.

\textbf{MVP Knowledge} 
To verify the effectiveness of the fine-tuning process, we choose MVP \cite{DBLP:conf/acl/TangLZW23} as the off-the-shelf KG-to-Text model. MVP is a text-generation LLM that is first pre-trained on 77 datasets over 11 diverse natural language generation (NLG) tasks in a supervised text-to-text format and then further pre-trained task-specific soft prompts to enhance the model's ability on specific tasks. We use MVP-data-to-text, a variant of MVP which is pre-trained on labeled data-to-text datasets to perform KG-to-Text in the zero-shot scenarios. We do not use this knowledge representation format for ZJQA since MVP does not support Chinese. 

\begin{table*}[h]
\caption{The detailed comparison results over no knowledge and triple knowledge. The below section of the table is the results of our framework implemented by different KG-to-Text models. We use ``Helpful'' and ``Harmful'' to evaluate the effectiveness of different knowledge representation formats and KG-to-Text methods. We adopt Llama-2-chat (13B) as the question-answering model. The best score is emphasized in bold.}
\begin{tabular}{l|cc|cc}
    \hline
   \multirow{2}{*}{Knowledge Format} & \multicolumn{2}{c|}{\textbf{No Knowledge Baseline}} & \multicolumn{2}{c}{\textbf{Triple Knowledge Baseline}}\\
    &\textbf{Helpful} $\uparrow$ &\textbf{Harmful} $\downarrow$ &\textbf{Helpful} $\uparrow$ & \textbf{Harmful} $\downarrow$\\
    \hline
    Triple Knowledge & 9,428 & 103 &- &-\\
    MTL Knowledge & 8,201 & 303 & 334 &1,761\\
    MVP Knowledge & 9,583 & 70 & 443 &255\\
    \hline
    \textbf{Flan-T5$_{(3B)}$}& 9,672 & 49 & 453 &155\\
    \textbf{Llama-2-chat$_{(7B)}$} & 9,690 & 46  &456 &137\\
    \textbf{Llama-2-chat$_{(13B)}$} & \textbf{9,703} & \textbf{38} &\textbf{459} &\textbf{119}\\
    \hline
\end{tabular}
\label{Table 3}
\end{table*}

\textbf{MTL Knowledge} 
We select MTL-data-to-text as another off-the-shelf KG-to-Text model. This model is a different variant of MVP and is pre-trained on a mixture of labeled data-to-text datasets. However, compared with MVP-data-to-text, it lacks training on other NLG tasks and task-specific soft prompts pre-training. We do not use this model for ZJQA due to its lack of support for Chinese.

Table \ref{Table 2} shows the experimental results of different knowledge representation formats based on various LLMs as the question-answering models. Compared with other knowledge representation formats, knowledge generated by our framework shows further improvement across multiple LLMs on KGQA.

\subsection{Analysis}

\textbf{Impact of Knowledge Representation Formats}
Table \ref{Table 2} demonstrates that different representations of the same retrieved triples have a significant influence on KGQA. The above section of Table \ref{Table 2} shows the results of the baseline representation formats. Without incorporating external knowledge, the performance of the question-answering model is the worst. This indicates that LLMs are incapable of storing all factual knowledge within their vast parameters, leading to issues of factual inaccuracy and knowledge missing. Other KG-augmented LLM methods outperform this baseline by a large margin, demonstrating the effectiveness of incorporating KG knowledge relevant to the questions. Among them, MTL knowledge exhibits the smallest improvement. This is because MTL is only pre-trained on multiple data-to-text datasets and its natural language understanding and generation capabilities are not strong enough. Therefore, the transformation from subgraphs to text loses semantic information and leads to limited enhancement. MVP-generated knowledge and triple-form knowledge result in comparable results in the question-answering task. The reason is that MVP is pre-trained on multiple NLG tasks and further fine-tuned on data-to-text datasets, possessing stronger capabilities in text comprehension and generation compared to MTL. Nevertheless, it lacks fine-tuning on domain-specific KG-to-Text corpus, leading to similar results as triplet knowledge. Triple-form knowledge is the most common approach in previous work. However, experimental results show that LLMs still struggle to extract semantics effectively from triples. This suggests that LLMs prefer receiving textual knowledge, as they have been pre-trained on massive corpora and structured triples are just a part of it. The below section of Table \ref{Table 2} shows the results of our framework. Our framework employs multiple KG-to-Text models, surpassing all baselines on various question-answering models. This not only demonstrates that our KG-to-Text method can generate answer-sensitive textual knowledge but also highlights the robust applicability of our framework to mainstream LLMs.

\textbf{Impact of LLMs}
The below section of Table \ref{Table 2} illustrates different performances of our proposed framework based on different LLMs. Generally, Llama-2-chat is better at KG-to-Text compared with Flan-T5-XL. We speculate this disparity can be attributed to differences in model parameters. The smaller size of Flan-T5-XL (3B) compared to Llama-2-chat (7B, 13B) may be responsible for the inferior performance. The two-parameter versions of Llama perform comparably. We believe this is due to the relative simplicity of KG-to-Text compared to other NLG tasks and 7B parameters are enough. For question-answering models, Llama-2-chat (13B) achieves the best performance on MetaQA while ChatGPT demonstrates the poorest performance. It is noteworthy that ChatGPT outperforms Llama-2-chat (7B, 13B) when no knowledge is provided. This observation indicates that ChatGPT has a greater capacity to retain knowledge due to its significantly larger parameter size. However, it does not leverage relevant knowledge as effectively as Llama-2-chat. For ZJQA, we employ the Chinese version of Llama-2-chat, Chinese-Alpaca-2. It is further trained based on Llama-2 using a substantial Chinese corpus and instruction data. Despite having significantly different model parameters, it achieves performance on ZJQA that matches or even surpasses ChatGPT. This emphasizes the significance of continued pre-training. Chinese-Alpaca-2 (13B) excels in almost all knowledge representation formats but lags behind ChatGPT in triple-form knowledge. Apart from differences in model parameters, we posit this is because ChatGPT has been trained on a wider range of structured data, enabling it to extract semantic information from structured data more effectively.

\textbf{Helpfulness/Harmfulness of Generated Knowledge} We further analyze the experimental results of MetaQA to investigate the positive and negative impact of knowledge generated by different methods on question-answering models. We establish two baselines: ``no knowledge'' and ``triple knowledge'', and compare other knowledge formats with these two baselines. We count the number of questions where the baseline answers incorrectly but other knowledge formats answer correctly (i.e., helpful), as well as the number of questions where the baseline answers correctly but other knowledge formats answer incorrectly (i.e., harmful). We choose Llama-2-chat (13B) as the question-answering model. The detailed information is presented in Table \ref{Table 3}.

The comparison with no knowledge baseline indicates that the knowledge representation format employed in our framework can better assist LLMs on KGQA and has fewer adverse effects. Besides, experimental results on the triple knowledge baseline reveal that our KG-to-Text method can generate the most informative textual statements for KGQA. This highlights the necessity of fine-tuning KG-to-Text models.

\section{Conclusion}
In this paper, we propose Retrieve-Rewrite-Answer, a KG-to-Text enhanced LLMs framework for KGQA. Our framework adopts an answer-sensitive KG-to-Text method to generate textual knowledge which is most informative for KGQA. To address the challenge of missing annotation data of KG-to-Text, we design a KG-to-Text corpus generation method based on the feedback of question-answering models. Experimental results demonstrate that our framework outperforms previous KG-augmented methods by a large margin. Besides, compared to other verbalization methods, the free-form knowledge text generated by our framework is more comprehensible to LLMs and achieves superior results on KGQA. However, this framework requires fine-tuning KG-to-Text models on graph-to-text corpus and does not incorporate extra data sources besides KG. In future work, we plan to explore the integration of additional knowledge resources and design an approach capable of generating question-related knowledge in zero-shot scenarios.
\bibliographystyle{ACM-Reference-Format}
\bibliography{sample-base}


\begin{thebibliography}{46}


\ifx \showCODEN    \undefined \def \showCODEN     #1{\unskip}     \fi
\ifx \showDOI      \undefined \def \showDOI       #1{#1}\fi
\ifx \showISBNx    \undefined \def \showISBNx     #1{\unskip}     \fi
\ifx \showISBNxiii \undefined \def \showISBNxiii  #1{\unskip}     \fi
\ifx \showISSN     \undefined \def \showISSN      #1{\unskip}     \fi
\ifx \showLCCN     \undefined \def \showLCCN      #1{\unskip}     \fi
\ifx \shownote     \undefined \def \shownote      #1{#1}          \fi
\ifx \showarticletitle \undefined \def \showarticletitle #1{#1}   \fi
\ifx \showURL      \undefined \def \showURL       {\relax}        \fi
\providecommand\bibfield[2]{#2}
\providecommand\bibinfo[2]{#2}
\providecommand\natexlab[1]{#1}
\providecommand\showeprint[2][]{arXiv:#2}

\bibitem[Baek et~al\mbox{.}(2023)]%
        {DBLP:journals/corr/abs-2306-04136}
\bibfield{author}{\bibinfo{person}{Jinheon Baek}, \bibinfo{person}{Alham~Fikri
  Aji}, {and} \bibinfo{person}{Amir Saffari}.} \bibinfo{year}{2023}\natexlab{}.
\newblock \showarticletitle{Knowledge-Augmented Language Model Prompting for
  Zero-Shot Knowledge Graph Question Answering}. In
  \bibinfo{booktitle}{\emph{Proceedings of the 1st Workshop on Natural Language
  Reasoning and Structured Explanations (NLRSE)}}.
  \bibinfo{publisher}{Association for Computational Linguistics},
  \bibinfo{address}{Toronto, Canada}, \bibinfo{pages}{78--106}.
\newblock
\urldef\tempurl%
\url{https://doi.org/10.18653/v1/2023.nlrse-1.7}
\showDOI{\tempurl}


\bibitem[Berant et~al\mbox{.}(2013)]%
        {DBLP:conf/emnlp/BerantCFL13}
\bibfield{author}{\bibinfo{person}{Jonathan Berant}, \bibinfo{person}{Andrew
  Chou}, \bibinfo{person}{Roy Frostig}, {and} \bibinfo{person}{Percy Liang}.}
  \bibinfo{year}{2013}\natexlab{}.
\newblock \showarticletitle{Semantic Parsing on Freebase from Question-Answer
  Pairs}. In \bibinfo{booktitle}{\emph{{EMNLP}}}. \bibinfo{publisher}{{ACL}},
  \bibinfo{pages}{1533--1544}.
\newblock


\bibitem[Cai and Lam(2020)]%
        {DBLP:conf/aaai/CaiL20}
\bibfield{author}{\bibinfo{person}{Deng Cai} {and} \bibinfo{person}{Wai Lam}.}
  \bibinfo{year}{2020}\natexlab{}.
\newblock \showarticletitle{Graph Transformer for Graph-to-Sequence Learning}.
  In \bibinfo{booktitle}{\emph{{AAAI}}}. \bibinfo{publisher}{{AAAI} Press},
  \bibinfo{pages}{7464--7471}.
\newblock


\bibitem[Chung et~al\mbox{.}(2022)]%
        {DBLP:journals/corr/abs-2210-11416}
\bibfield{author}{\bibinfo{person}{Hyung~Won Chung}, \bibinfo{person}{Le Hou},
  \bibinfo{person}{Shayne Longpre}, \bibinfo{person}{Barret Zoph},
  \bibinfo{person}{Yi Tay}, \bibinfo{person}{William Fedus},
  \bibinfo{person}{Yunxuan Li}, \bibinfo{person}{Xuezhi Wang},
  \bibinfo{person}{Mostafa Dehghani}, \bibinfo{person}{Siddhartha Brahma},
  {et~al\mbox{.}}} \bibinfo{year}{2022}\natexlab{}.
\newblock \showarticletitle{Scaling Instruction-Finetuned Language Models}.
\newblock  (\bibinfo{year}{2022}).
\newblock
\urldef\tempurl%
\url{https://doi.org/10.48550/arXiv.2210.11416}
\showDOI{\tempurl}
\showeprint{2210.11416}


\bibitem[Dai et~al\mbox{.}(2023)]%
        {dai2023auggpt}
\bibfield{author}{\bibinfo{person}{Haixing Dai}, \bibinfo{person}{Zhengliang
  Liu}, \bibinfo{person}{Wenxiong Liao}, \bibinfo{person}{Xiaoke Huang},
  \bibinfo{person}{Yihan Cao}, \bibinfo{person}{Zihao Wu}, \bibinfo{person}{Lin
  Zhao}, \bibinfo{person}{Shaochen Xu}, \bibinfo{person}{Wei Liu},
  \bibinfo{person}{Ninghao Liu}, {et~al\mbox{.}}}
  \bibinfo{year}{2023}\natexlab{}.
\newblock \showarticletitle{AugGPT: Leveraging ChatGPT for Text Data
  Augmentation}.
\newblock  (\bibinfo{year}{2023}).
\newblock
\urldef\tempurl%
\url{https://doi.org/10.48550/arXiv.2302.13007}
\showDOI{\tempurl}
\showeprint{2302.13007}


\bibitem[Guo et~al\mbox{.}(2019)]%
        {DBLP:journals/tacl/GuoZTL19}
\bibfield{author}{\bibinfo{person}{Zhijiang Guo}, \bibinfo{person}{Yan Zhang},
  \bibinfo{person}{Zhiyang Teng}, {and} \bibinfo{person}{Wei Lu}.}
  \bibinfo{year}{2019}\natexlab{}.
\newblock \showarticletitle{Densely Connected Graph Convolutional Networks for
  Graph-to-Sequence Learning}.
\newblock \bibinfo{journal}{\emph{Trans. Assoc. Comput. Linguistics}}
  \bibinfo{volume}{7} (\bibinfo{year}{2019}), \bibinfo{pages}{297--312}.
\newblock


\bibitem[Han and Shareghi(2022)]%
        {DBLP:conf/emnlp/HanS22}
\bibfield{author}{\bibinfo{person}{Jiuzhou Han} {and} \bibinfo{person}{Ehsan
  Shareghi}.} \bibinfo{year}{2022}\natexlab{}.
\newblock \showarticletitle{Self-supervised Graph Masking Pre-training for
  Graph-to-Text Generation}. In \bibinfo{booktitle}{\emph{{EMNLP}}}.
  \bibinfo{publisher}{Association for Computational Linguistics},
  \bibinfo{pages}{4845--4853}.
\newblock


\bibitem[Hu et~al\mbox{.}(2022)]%
        {DBLP:conf/iclr/HuSWALWWC22}
\bibfield{author}{\bibinfo{person}{Edward~J. Hu}, \bibinfo{person}{Yelong
  Shen}, \bibinfo{person}{Phillip Wallis}, \bibinfo{person}{Zeyuan
  Allen{-}Zhu}, \bibinfo{person}{Yuanzhi Li}, \bibinfo{person}{Shean Wang},
  \bibinfo{person}{Lu Wang}, {and} \bibinfo{person}{Weizhu Chen}.}
  \bibinfo{year}{2022}\natexlab{}.
\newblock \showarticletitle{LoRA: Low-Rank Adaptation of Large Language
  Models}. In \bibinfo{booktitle}{\emph{{ICLR}}}.
  \bibinfo{publisher}{OpenReview.net}.
\newblock


\bibitem[Hu et~al\mbox{.}(2023)]%
        {hu2023empirical}
\bibfield{author}{\bibinfo{person}{Nan Hu}, \bibinfo{person}{Yike Wu},
  \bibinfo{person}{Guilin Qi}, \bibinfo{person}{Dehai Min},
  \bibinfo{person}{Jiaoyan Chen}, \bibinfo{person}{Jeff~Z Pan}, {and}
  \bibinfo{person}{Zafar Ali}.} \bibinfo{year}{2023}\natexlab{}.
\newblock \showarticletitle{An empirical study of pre-trained language models
  in simple knowledge graph question answering}.
\newblock \bibinfo{journal}{\emph{World Wide Web}} (\bibinfo{year}{2023}),
  \bibinfo{pages}{1--32}.
\newblock


\bibitem[Huang et~al\mbox{.}(2023)]%
        {DBLP:journals/corr/abs-2305-15062}
\bibfield{author}{\bibinfo{person}{Quzhe Huang}, \bibinfo{person}{Mingxu Tao},
  \bibinfo{person}{Zhenwei An}, \bibinfo{person}{Chen Zhang},
  \bibinfo{person}{Cong Jiang}, \bibinfo{person}{Zhibin Chen},
  \bibinfo{person}{Zirui Wu}, {and} \bibinfo{person}{Yansong Feng}.}
  \bibinfo{year}{2023}\natexlab{}.
\newblock \showarticletitle{Lawyer LLaMA Technical Report}.
\newblock  (\bibinfo{year}{2023}).
\newblock
\urldef\tempurl%
\url{https://doi.org/10.48550/arXiv.2305.15062}
\showDOI{\tempurl}
\showeprint{2305.15062}


\bibitem[IV et~al\mbox{.}(2019)]%
        {DBLP:conf/acl/LoganLPGS19}
\bibfield{author}{\bibinfo{person}{Robert L.~Logan IV},
  \bibinfo{person}{Nelson~F. Liu}, \bibinfo{person}{Matthew~E. Peters},
  \bibinfo{person}{Matt Gardner}, {and} \bibinfo{person}{Sameer Singh}.}
  \bibinfo{year}{2019}\natexlab{}.
\newblock \showarticletitle{Barack's Wife Hillary: Using Knowledge Graphs for
  Fact-Aware Language Modeling}. In \bibinfo{booktitle}{\emph{{ACL} {(1)}}}.
  \bibinfo{publisher}{Association for Computational Linguistics},
  \bibinfo{pages}{5962--5971}.
\newblock


\bibitem[Ji et~al\mbox{.}(2023)]%
        {DBLP:journals/csur/JiLFYSXIBMF23}
\bibfield{author}{\bibinfo{person}{Ziwei Ji}, \bibinfo{person}{Nayeon Lee},
  \bibinfo{person}{Rita Frieske}, \bibinfo{person}{Tiezheng Yu},
  \bibinfo{person}{Dan Su}, \bibinfo{person}{Yan Xu}, \bibinfo{person}{Etsuko
  Ishii}, \bibinfo{person}{Yejin Bang}, \bibinfo{person}{Andrea Madotto}, {and}
  \bibinfo{person}{Pascale Fung}.} \bibinfo{year}{2023}\natexlab{}.
\newblock \showarticletitle{Survey of Hallucination in Natural Language
  Generation}.
\newblock \bibinfo{journal}{\emph{{ACM} Comput. Surv.}} \bibinfo{volume}{55},
  \bibinfo{number}{12} (\bibinfo{year}{2023}), \bibinfo{pages}{248:1--248:38}.
\newblock


\bibitem[Ke et~al\mbox{.}(2021)]%
        {DBLP:conf/acl/KeJRCWSZH21}
\bibfield{author}{\bibinfo{person}{Pei Ke}, \bibinfo{person}{Haozhe Ji},
  \bibinfo{person}{Yu Ran}, \bibinfo{person}{Xin Cui}, \bibinfo{person}{Liwei
  Wang}, \bibinfo{person}{Linfeng Song}, \bibinfo{person}{Xiaoyan Zhu}, {and}
  \bibinfo{person}{Minlie Huang}.} \bibinfo{year}{2021}\natexlab{}.
\newblock \showarticletitle{JointGT: Graph-Text Joint Representation Learning
  for Text Generation from Knowledge Graphs}. In
  \bibinfo{booktitle}{\emph{{ACL/IJCNLP} (Findings)}}
  \emph{(\bibinfo{series}{Findings of {ACL}},
  Vol.~\bibinfo{volume}{{ACL/IJCNLP} 2021})}. \bibinfo{publisher}{Association
  for Computational Linguistics}, \bibinfo{pages}{2526--2538}.
\newblock


\bibitem[Komeili et~al\mbox{.}(2022)]%
        {komeili-etal-2022-internet}
\bibfield{author}{\bibinfo{person}{Mojtaba Komeili}, \bibinfo{person}{Kurt
  Shuster}, {and} \bibinfo{person}{Jason Weston}.}
  \bibinfo{year}{2022}\natexlab{}.
\newblock \showarticletitle{{I}nternet-Augmented Dialogue Generation}. In
  \bibinfo{booktitle}{\emph{Proceedings of the 60th Annual Meeting of the
  Association for Computational Linguistics (Volume 1: Long Papers)}}.
  \bibinfo{publisher}{Association for Computational Linguistics},
  \bibinfo{address}{Dublin, Ireland}, \bibinfo{pages}{8460--8478}.
\newblock
\urldef\tempurl%
\url{https://doi.org/10.18653/v1/2022.acl-long.579}
\showDOI{\tempurl}


\bibitem[Koncel{-}Kedziorski et~al\mbox{.}(2019)]%
        {DBLP:conf/naacl/Koncel-Kedziorski19}
\bibfield{author}{\bibinfo{person}{Rik Koncel{-}Kedziorski},
  \bibinfo{person}{Dhanush Bekal}, \bibinfo{person}{Yi Luan},
  \bibinfo{person}{Mirella Lapata}, {and} \bibinfo{person}{Hannaneh
  Hajishirzi}.} \bibinfo{year}{2019}\natexlab{}.
\newblock \showarticletitle{Text Generation from Knowledge Graphs with Graph
  Transformers}. In \bibinfo{booktitle}{\emph{{NAACL-HLT} {(1)}}}.
  \bibinfo{publisher}{Association for Computational Linguistics},
  \bibinfo{pages}{2284--2293}.
\newblock


\bibitem[Laskar et~al\mbox{.}(2023)]%
        {DBLP:journals/corr/abs-2305-06147}
\bibfield{author}{\bibinfo{person}{Md~Tahmid~Rahman Laskar},
  \bibinfo{person}{Mizanur Rahman}, \bibinfo{person}{Israt Jahan},
  \bibinfo{person}{Enamul Hoque}, {and} \bibinfo{person}{Jimmy Huang}.}
  \bibinfo{year}{2023}\natexlab{}.
\newblock \showarticletitle{CQSumDP: A ChatGPT-Annotated Resource for
  Query-Focused Abstractive Summarization Based on Debatepedia}.
\newblock  (\bibinfo{year}{2023}).
\newblock
\urldef\tempurl%
\url{https://doi.org/10.48550/arXiv.2305.06147}
\showDOI{\tempurl}
\showeprint{2305.06147}


\bibitem[Lewis et~al\mbox{.}(2020)]%
        {DBLP:conf/acl/LewisLGGMLSZ20}
\bibfield{author}{\bibinfo{person}{Mike Lewis}, \bibinfo{person}{Yinhan Liu},
  \bibinfo{person}{Naman Goyal}, \bibinfo{person}{Marjan Ghazvininejad},
  \bibinfo{person}{Abdelrahman Mohamed}, \bibinfo{person}{Omer Levy},
  \bibinfo{person}{Veselin Stoyanov}, {and} \bibinfo{person}{Luke
  Zettlemoyer}.} \bibinfo{year}{2020}\natexlab{}.
\newblock \showarticletitle{{BART:} Denoising Sequence-to-Sequence Pre-training
  for Natural Language Generation, Translation, and Comprehension}. In
  \bibinfo{booktitle}{\emph{{ACL}}}. \bibinfo{publisher}{Association for
  Computational Linguistics}, \bibinfo{pages}{7871--7880}.
\newblock


\bibitem[Liu et~al\mbox{.}(2022)]%
        {DBLP:conf/emnlp/0010HLHWHC22}
\bibfield{author}{\bibinfo{person}{Jiacheng Liu}, \bibinfo{person}{Skyler
  Hallinan}, \bibinfo{person}{Ximing Lu}, \bibinfo{person}{Pengfei He},
  \bibinfo{person}{Sean Welleck}, \bibinfo{person}{Hannaneh Hajishirzi}, {and}
  \bibinfo{person}{Yejin Choi}.} \bibinfo{year}{2022}\natexlab{}.
\newblock \showarticletitle{Rainier: Reinforced Knowledge Introspector for
  Commonsense Question Answering}. In \bibinfo{booktitle}{\emph{{EMNLP}}}.
  \bibinfo{publisher}{Association for Computational Linguistics},
  \bibinfo{pages}{8938--8958}.
\newblock


\bibitem[Lu et~al\mbox{.}(2023)]%
        {DBLP:journals/corr/abs-2302-09432}
\bibfield{author}{\bibinfo{person}{Dakuan Lu}, \bibinfo{person}{Hengkui Wu},
  \bibinfo{person}{Jiaqing Liang}, \bibinfo{person}{Yipei Xu},
  \bibinfo{person}{Qianyu He}, \bibinfo{person}{Yipeng Geng},
  \bibinfo{person}{Mengkun Han}, \bibinfo{person}{Yingsi Xin}, {and}
  \bibinfo{person}{Yanghua Xiao}.} \bibinfo{year}{2023}\natexlab{}.
\newblock \showarticletitle{BBT-Fin: Comprehensive Construction of Chinese
  Financial Domain Pre-trained Language Model, Corpus and Benchmark}.
\newblock  (\bibinfo{year}{2023}).
\newblock
\urldef\tempurl%
\url{https://doi.org/10.48550/arXiv.2302.09432}
\showDOI{\tempurl}
\showeprint{2302.09432}


\bibitem[Mallen et~al\mbox{.}(2023)]%
        {DBLP:conf/acl/MallenAZDKH23}
\bibfield{author}{\bibinfo{person}{Alex Mallen}, \bibinfo{person}{Akari Asai},
  \bibinfo{person}{Victor Zhong}, \bibinfo{person}{Rajarshi Das},
  \bibinfo{person}{Daniel Khashabi}, {and} \bibinfo{person}{Hannaneh
  Hajishirzi}.} \bibinfo{year}{2023}\natexlab{}.
\newblock \showarticletitle{When Not to Trust Language Models: Investigating
  Effectiveness of Parametric and Non-Parametric Memories}. In
  \bibinfo{booktitle}{\emph{{ACL} {(1)}}}. \bibinfo{publisher}{Association for
  Computational Linguistics}, \bibinfo{pages}{9802--9822}.
\newblock


\bibitem[Marcheggiani and Perez{-}Beltrachini(2018)]%
        {DBLP:conf/inlg/MarcheggianiP18}
\bibfield{author}{\bibinfo{person}{Diego Marcheggiani} {and}
  \bibinfo{person}{Laura Perez{-}Beltrachini}.}
  \bibinfo{year}{2018}\natexlab{}.
\newblock \showarticletitle{Deep Graph Convolutional Encoders for Structured
  Data to Text Generation}. In \bibinfo{booktitle}{\emph{{INLG}}}.
  \bibinfo{publisher}{Association for Computational Linguistics},
  \bibinfo{pages}{1--9}.
\newblock


\bibitem[Ostyakova et~al\mbox{.}(2023)]%
        {ostyakova2023linguistic}
\bibfield{author}{\bibinfo{person}{Lidiia Ostyakova}, \bibinfo{person}{Kseniia
  PetukhovaO}, \bibinfo{person}{Veronika Smilga}, {and}
  \bibinfo{person}{Dilyara ZharikovaO}.} \bibinfo{year}{2023}\natexlab{}.
\newblock \showarticletitle{Linguistic Annotation Generation with ChatGPT: a
  Synthetic Dataset of Speech Functions for Discourse Annotation of Casual
  Conversations}. In \bibinfo{booktitle}{\emph{Proceedings of the International
  Conference “Dialogue}}, Vol.~\bibinfo{volume}{2023}.
\newblock


\bibitem[Radford et~al\mbox{.}(2019)]%
        {radfordlanguage}
\bibfield{author}{\bibinfo{person}{Alec Radford}, \bibinfo{person}{Jeffrey Wu},
  \bibinfo{person}{Rewon Child}, \bibinfo{person}{David Luan},
  \bibinfo{person}{Dario Amodei}, {and} \bibinfo{person}{Ilya Sutskever}.}
  \bibinfo{year}{2019}\natexlab{}.
\newblock \bibinfo{booktitle}{\emph{Language Models are Unsupervised Multitask
  Learners}}.
\newblock \bibinfo{type}{{T}echnical {R}eport}.
\newblock


\bibitem[Raffel et~al\mbox{.}(2020)]%
        {DBLP:journals/jmlr/RaffelSRLNMZLL20}
\bibfield{author}{\bibinfo{person}{Colin Raffel}, \bibinfo{person}{Noam
  Shazeer}, \bibinfo{person}{Adam Roberts}, \bibinfo{person}{Katherine Lee},
  \bibinfo{person}{Sharan Narang}, \bibinfo{person}{Michael Matena},
  \bibinfo{person}{Yanqi Zhou}, \bibinfo{person}{Wei Li}, {and}
  \bibinfo{person}{Peter~J. Liu}.} \bibinfo{year}{2020}\natexlab{}.
\newblock \showarticletitle{Exploring the Limits of Transfer Learning with a
  Unified Text-to-Text Transformer}.
\newblock \bibinfo{journal}{\emph{J. Mach. Learn. Res.}}  \bibinfo{volume}{21}
  (\bibinfo{year}{2020}), \bibinfo{pages}{140:1--140:67}.
\newblock


\bibitem[Ribeiro et~al\mbox{.}(2021)]%
        {ribeiro-etal-2021-investigating}
\bibfield{author}{\bibinfo{person}{Leonardo F.~R. Ribeiro},
  \bibinfo{person}{Martin Schmitt}, \bibinfo{person}{Hinrich Sch{\"u}tze},
  {and} \bibinfo{person}{Iryna Gurevych}.} \bibinfo{year}{2021}\natexlab{}.
\newblock \showarticletitle{Investigating Pretrained Language Models for
  Graph-to-Text Generation}. In \bibinfo{booktitle}{\emph{Proceedings of the
  3rd Workshop on Natural Language Processing for Conversational AI}}.
  \bibinfo{publisher}{Association for Computational Linguistics},
  \bibinfo{address}{Online}, \bibinfo{pages}{211--227}.
\newblock
\urldef\tempurl%
\url{https://doi.org/10.18653/v1/2021.nlp4convai-1.20}
\showDOI{\tempurl}


\bibitem[Rohrbach et~al\mbox{.}(2018)]%
        {DBLP:conf/emnlp/RohrbachHBDS18}
\bibfield{author}{\bibinfo{person}{Anna Rohrbach}, \bibinfo{person}{Lisa~Anne
  Hendricks}, \bibinfo{person}{Kaylee Burns}, \bibinfo{person}{Trevor Darrell},
  {and} \bibinfo{person}{Kate Saenko}.} \bibinfo{year}{2018}\natexlab{}.
\newblock \showarticletitle{Object Hallucination in Image Captioning}. In
  \bibinfo{booktitle}{\emph{{EMNLP}}}. \bibinfo{publisher}{Association for
  Computational Linguistics}, \bibinfo{pages}{4035--4045}.
\newblock


\bibitem[Saffari et~al\mbox{.}(2021)]%
        {DBLP:conf/emnlp/SaffariOSA21}
\bibfield{author}{\bibinfo{person}{Amir Saffari}, \bibinfo{person}{Armin
  Oliya}, \bibinfo{person}{Priyanka Sen}, {and} \bibinfo{person}{Tom Ayoola}.}
  \bibinfo{year}{2021}\natexlab{}.
\newblock \showarticletitle{End-to-End Entity Resolution and Question Answering
  Using Differentiable Knowledge Graphs}. In \bibinfo{booktitle}{\emph{{EMNLP}
  {(1)}}}. \bibinfo{publisher}{Association for Computational Linguistics},
  \bibinfo{pages}{4193--4200}.
\newblock


\bibitem[Sanh et~al\mbox{.}(2022)]%
        {DBLP:conf/iclr/SanhWRBSACSRDBX22}
\bibfield{author}{\bibinfo{person}{Victor Sanh}, \bibinfo{person}{Albert
  Webson}, \bibinfo{person}{Colin Raffel}, \bibinfo{person}{Stephen~H. Bach},
  \bibinfo{person}{Lintang Sutawika}, \bibinfo{person}{Zaid Alyafeai},
  \bibinfo{person}{Antoine Chaffin}, \bibinfo{person}{Arnaud Stiegler},
  \bibinfo{person}{Arun Raja}, \bibinfo{person}{Manan Dey}, {et~al\mbox{.}}}
  \bibinfo{year}{2022}\natexlab{}.
\newblock \showarticletitle{Multitask Prompted Training Enables Zero-Shot Task
  Generalization}. In \bibinfo{booktitle}{\emph{{ICLR}}}.
  \bibinfo{publisher}{OpenReview.net}.
\newblock


\bibitem[Saxena et~al\mbox{.}(2022)]%
        {DBLP:conf/acl/SaxenaKG22}
\bibfield{author}{\bibinfo{person}{Apoorv Saxena}, \bibinfo{person}{Adrian
  Kochsiek}, {and} \bibinfo{person}{Rainer Gemulla}.}
  \bibinfo{year}{2022}\natexlab{}.
\newblock \showarticletitle{Sequence-to-Sequence Knowledge Graph Completion and
  Question Answering}. In \bibinfo{booktitle}{\emph{{ACL} {(1)}}}.
  \bibinfo{publisher}{Association for Computational Linguistics},
  \bibinfo{pages}{2814--2828}.
\newblock


\bibitem[Saxena et~al\mbox{.}(2020)]%
        {DBLP:conf/acl/SaxenaTT20}
\bibfield{author}{\bibinfo{person}{Apoorv Saxena}, \bibinfo{person}{Aditay
  Tripathi}, {and} \bibinfo{person}{Partha~P. Talukdar}.}
  \bibinfo{year}{2020}\natexlab{}.
\newblock \showarticletitle{Improving Multi-hop Question Answering over
  Knowledge Graphs using Knowledge Base Embeddings}. In
  \bibinfo{booktitle}{\emph{{ACL}}}. \bibinfo{publisher}{Association for
  Computational Linguistics}, \bibinfo{pages}{4498--4507}.
\newblock


\bibitem[Sen et~al\mbox{.}(2022)]%
        {DBLP:conf/coling/SenAS22}
\bibfield{author}{\bibinfo{person}{Priyanka Sen}, \bibinfo{person}{Alham~Fikri
  Aji}, {and} \bibinfo{person}{Amir Saffari}.} \bibinfo{year}{2022}\natexlab{}.
\newblock \showarticletitle{Mintaka: {A} Complex, Natural, and Multilingual
  Dataset for End-to-End Question Answering}. In
  \bibinfo{booktitle}{\emph{{COLING}}}. \bibinfo{publisher}{International
  Committee on Computational Linguistics}, \bibinfo{pages}{1604--1619}.
\newblock


\bibitem[Sen et~al\mbox{.}(2023)]%
        {sen-etal-2023-knowledge}
\bibfield{author}{\bibinfo{person}{Priyanka Sen}, \bibinfo{person}{Sandeep
  Mavadia}, {and} \bibinfo{person}{Amir Saffari}.}
  \bibinfo{year}{2023}\natexlab{}.
\newblock \showarticletitle{Knowledge Graph-augmented Language Models for
  Complex Question Answering}. In \bibinfo{booktitle}{\emph{Proceedings of the
  1st Workshop on Natural Language Reasoning and Structured Explanations
  (NLRSE)}}. \bibinfo{publisher}{Association for Computational Linguistics},
  \bibinfo{address}{Toronto, Canada}, \bibinfo{pages}{1--8}.
\newblock
\urldef\tempurl%
\url{https://doi.org/10.18653/v1/2023.nlrse-1.1}
\showDOI{\tempurl}


\bibitem[Talmor and Berant(2018)]%
        {DBLP:conf/naacl/TalmorB18}
\bibfield{author}{\bibinfo{person}{Alon Talmor} {and} \bibinfo{person}{Jonathan
  Berant}.} \bibinfo{year}{2018}\natexlab{}.
\newblock \showarticletitle{The Web as a Knowledge-Base for Answering Complex
  Questions}. In \bibinfo{booktitle}{\emph{{NAACL-HLT}}}.
  \bibinfo{publisher}{Association for Computational Linguistics},
  \bibinfo{pages}{641--651}.
\newblock


\bibitem[Tan et~al\mbox{.}(2023)]%
        {DBLP:journals/corr/abs-2303-07992}
\bibfield{author}{\bibinfo{person}{Yiming Tan}, \bibinfo{person}{Dehai Min},
  \bibinfo{person}{Yu Li}, \bibinfo{person}{Wenbo Li}, \bibinfo{person}{Nan
  Hu}, \bibinfo{person}{Yongrui Chen}, {and} \bibinfo{person}{Guilin Qi}.}
  \bibinfo{year}{2023}\natexlab{}.
\newblock \showarticletitle{Evaluation of ChatGPT as a Question Answering
  System for Answering Complex Questions}.
\newblock  (\bibinfo{year}{2023}).
\newblock
\urldef\tempurl%
\url{https://doi.org/10.48550/arXiv.2303.07992}
\showDOI{\tempurl}
\showeprint{2303.07992}


\bibitem[Tang et~al\mbox{.}(2023)]%
        {DBLP:conf/acl/TangLZW23}
\bibfield{author}{\bibinfo{person}{Tianyi Tang}, \bibinfo{person}{Junyi Li},
  \bibinfo{person}{Wayne~Xin Zhao}, {and} \bibinfo{person}{Ji{-}Rong Wen}.}
  \bibinfo{year}{2023}\natexlab{}.
\newblock \showarticletitle{{MVP:} Multi-task Supervised Pre-training for
  Natural Language Generation}. In \bibinfo{booktitle}{\emph{{ACL}
  (Findings)}}. \bibinfo{publisher}{Association for Computational Linguistics},
  \bibinfo{pages}{8758--8794}.
\newblock


\bibitem[Touvron et~al\mbox{.}(2023a)]%
        {DBLP:journals/corr/abs-2302-13971}
\bibfield{author}{\bibinfo{person}{Hugo Touvron}, \bibinfo{person}{Thibaut
  Lavril}, \bibinfo{person}{Gautier Izacard}, \bibinfo{person}{Xavier
  Martinet}, \bibinfo{person}{Marie-Anne Lachaux}, \bibinfo{person}{Timothée
  Lacroix}, \bibinfo{person}{Baptiste Rozière}, \bibinfo{person}{Naman Goyal},
  \bibinfo{person}{Eric Hambro}, \bibinfo{person}{Faisal Azhar},
  {et~al\mbox{.}}} \bibinfo{year}{2023}\natexlab{a}.
\newblock \showarticletitle{LLaMA: Open and Efficient Foundation Language
  Models}.
\newblock  (\bibinfo{year}{2023}).
\newblock
\urldef\tempurl%
\url{https://doi.org/10.48550/arXiv.2302.13971}
\showDOI{\tempurl}
\showeprint{2302.13971}


\bibitem[Touvron et~al\mbox{.}(2023b)]%
        {DBLP:journals/corr/abs-2307-09288}
\bibfield{author}{\bibinfo{person}{Hugo Touvron}, \bibinfo{person}{Louis
  Martin}, \bibinfo{person}{Kevin Stone}, \bibinfo{person}{Peter Albert},
  \bibinfo{person}{Amjad Almahairi}, \bibinfo{person}{Yasmine Babaei},
  \bibinfo{person}{Nikolay Bashlykov}, \bibinfo{person}{Soumya Batra},
  \bibinfo{person}{Prajjwal Bhargava}, \bibinfo{person}{Shruti Bhosale},
  {et~al\mbox{.}}} \bibinfo{year}{2023}\natexlab{b}.
\newblock \showarticletitle{Llama 2: Open Foundation and Fine-Tuned Chat
  Models}.
\newblock  (\bibinfo{year}{2023}).
\newblock
\urldef\tempurl%
\url{https://doi.org/10.48550/arXiv.2307.09288}
\showDOI{\tempurl}
\showeprint{2307.09288}


\bibitem[Tu et~al\mbox{.}(2023)]%
        {DBLP:journals/corr/abs-2307-14334}
\bibfield{author}{\bibinfo{person}{Tao Tu}, \bibinfo{person}{Shekoofeh Azizi},
  \bibinfo{person}{Danny Driess}, \bibinfo{person}{Mike Schaekermann},
  \bibinfo{person}{Mohamed Amin}, \bibinfo{person}{Pi-Chuan Chang},
  \bibinfo{person}{Andrew Carroll}, \bibinfo{person}{Chuck Lau},
  \bibinfo{person}{Ryutaro Tanno}, \bibinfo{person}{Ira Ktena},
  {et~al\mbox{.}}} \bibinfo{year}{2023}\natexlab{}.
\newblock \showarticletitle{Towards Generalist Biomedical AI}.
\newblock  (\bibinfo{year}{2023}).
\newblock
\urldef\tempurl%
\url{https://doi.org/10.48550/arXiv.2307.14334}
\showDOI{\tempurl}
\showeprint{2307.14334}


\bibitem[Vaswani et~al\mbox{.}(2017)]%
        {DBLP:conf/nips/VaswaniSPUJGKP17}
\bibfield{author}{\bibinfo{person}{Ashish Vaswani}, \bibinfo{person}{Noam
  Shazeer}, \bibinfo{person}{Niki Parmar}, \bibinfo{person}{Jakob Uszkoreit},
  \bibinfo{person}{Llion Jones}, \bibinfo{person}{Aidan~N. Gomez},
  \bibinfo{person}{Lukasz Kaiser}, {and} \bibinfo{person}{Illia Polosukhin}.}
  \bibinfo{year}{2017}\natexlab{}.
\newblock \showarticletitle{Attention is All you Need}. In
  \bibinfo{booktitle}{\emph{{NIPS}}}. \bibinfo{pages}{5998--6008}.
\newblock


\bibitem[Velickovic et~al\mbox{.}(2018)]%
        {DBLP:conf/iclr/VelickovicCCRLB18}
\bibfield{author}{\bibinfo{person}{Petar Velickovic}, \bibinfo{person}{Guillem
  Cucurull}, \bibinfo{person}{Arantxa Casanova}, \bibinfo{person}{Adriana
  Romero}, \bibinfo{person}{Pietro Li{\`{o}}}, {and} \bibinfo{person}{Yoshua
  Bengio}.} \bibinfo{year}{2018}\natexlab{}.
\newblock \showarticletitle{Graph Attention Networks}. In
  \bibinfo{booktitle}{\emph{{ICLR} (Poster)}}.
  \bibinfo{publisher}{OpenReview.net}.
\newblock


\bibitem[Wang et~al\mbox{.}(2023)]%
        {DBLP:journals/corr/abs-2304-06975}
\bibfield{author}{\bibinfo{person}{Haochun Wang}, \bibinfo{person}{Chi Liu},
  \bibinfo{person}{Nuwa Xi}, \bibinfo{person}{Zewen Qiang},
  \bibinfo{person}{Sendong Zhao}, \bibinfo{person}{Bing Qin}, {and}
  \bibinfo{person}{Ting Liu}.} \bibinfo{year}{2023}\natexlab{}.
\newblock \showarticletitle{HuaTuo: Tuning LLaMA Model with Chinese Medical
  Knowledge}.
\newblock  (\bibinfo{year}{2023}).
\newblock
\urldef\tempurl%
\url{https://doi.org/10.48550/arXiv.2304.06975}
\showDOI{\tempurl}
\showeprint{2304.06975}


\bibitem[Yih et~al\mbox{.}(2016)]%
        {DBLP:conf/acl/YihRMCS16}
\bibfield{author}{\bibinfo{person}{Wen{-}tau Yih}, \bibinfo{person}{Matthew
  Richardson}, \bibinfo{person}{Christopher Meek}, \bibinfo{person}{Ming{-}Wei
  Chang}, {and} \bibinfo{person}{Jina Suh}.} \bibinfo{year}{2016}\natexlab{}.
\newblock \showarticletitle{The Value of Semantic Parse Labeling for Knowledge
  Base Question Answering}. In \bibinfo{booktitle}{\emph{{ACL} {(2)}}}.
  \bibinfo{publisher}{The Association for Computer Linguistics}.
\newblock


\bibitem[Yin et~al\mbox{.}(2016)]%
        {DBLP:conf/ijcai/YinJLSLL16}
\bibfield{author}{\bibinfo{person}{Jun Yin}, \bibinfo{person}{Xin Jiang},
  \bibinfo{person}{Zhengdong Lu}, \bibinfo{person}{Lifeng Shang},
  \bibinfo{person}{Hang Li}, {and} \bibinfo{person}{Xiaoming Li}.}
  \bibinfo{year}{2016}\natexlab{}.
\newblock \showarticletitle{Neural Generative Question Answering}. In
  \bibinfo{booktitle}{\emph{{IJCAI}}}. \bibinfo{publisher}{{IJCAI/AAAI} Press},
  \bibinfo{pages}{2972--2978}.
\newblock


\bibitem[Zhang et~al\mbox{.}(2018)]%
        {DBLP:conf/aaai/ZhangDKSS18}
\bibfield{author}{\bibinfo{person}{Yuyu Zhang}, \bibinfo{person}{Hanjun Dai},
  \bibinfo{person}{Zornitsa Kozareva}, \bibinfo{person}{Alexander~J. Smola},
  {and} \bibinfo{person}{Le Song}.} \bibinfo{year}{2018}\natexlab{}.
\newblock \showarticletitle{Variational Reasoning for Question Answering With
  Knowledge Graph}. In \bibinfo{booktitle}{\emph{{AAAI}}}.
  \bibinfo{publisher}{{AAAI} Press}, \bibinfo{pages}{6069--6076}.
\newblock


\bibitem[Zhou et~al\mbox{.}(2018)]%
        {DBLP:conf/coling/ZhouHZ18}
\bibfield{author}{\bibinfo{person}{Mantong Zhou}, \bibinfo{person}{Minlie
  Huang}, {and} \bibinfo{person}{Xiaoyan Zhu}.}
  \bibinfo{year}{2018}\natexlab{}.
\newblock \showarticletitle{An Interpretable Reasoning Network for
  Multi-Relation Question Answering}. In \bibinfo{booktitle}{\emph{{COLING}}}.
  \bibinfo{publisher}{Association for Computational Linguistics},
  \bibinfo{pages}{2010--2022}.
\newblock


\bibitem[Zhu et~al\mbox{.}(2019)]%
        {DBLP:conf/emnlp/ZhuLZQZZ19}
\bibfield{author}{\bibinfo{person}{Jie Zhu}, \bibinfo{person}{Junhui Li},
  \bibinfo{person}{Muhua Zhu}, \bibinfo{person}{Longhua Qian},
  \bibinfo{person}{Min Zhang}, {and} \bibinfo{person}{Guodong Zhou}.}
  \bibinfo{year}{2019}\natexlab{}.
\newblock \showarticletitle{Modeling Graph Structure in Transformer for Better
  AMR-to-Text Generation}. In \bibinfo{booktitle}{\emph{{EMNLP/IJCNLP} {(1)}}}.
  \bibinfo{publisher}{Association for Computational Linguistics},
  \bibinfo{pages}{5458--5467}.
\newblock


\end{thebibliography}

\end{CJK}

\end{document}